\documentclass[Numbered,sn-basic]{sn-jnl}

\usepackage{graphicx}%
\usepackage{multirow}%
\usepackage{amsmath,amssymb,amsfonts}%
\usepackage{amsthm}%
\usepackage{mathrsfs}%
\usepackage[title]{appendix}%
\usepackage{xcolor}%
\usepackage{textcomp}%
\usepackage{manyfoot}%
\usepackage{booktabs}%
\usepackage{algorithm}%
\usepackage{algorithmicx}%
\usepackage{algpseudocode}%
\usepackage{listings}%
\usepackage{tabu}                     
\usepackage{multicol}                 
\usepackage{float}                    
\usepackage{makecell}                 
\usepackage{picinpar}
\usepackage{hhline}
\usepackage{url}
\usepackage{xcolor}
\definecolor{newcolor}{rgb}{.8,.349,.1}
\usepackage{hyperref}
\usepackage[switch,pagewise]{lineno} 
\usepackage{natbib}
\usepackage{booktabs} 

\theoremstyle{thmstyleone}%
\theoremstyle{thmstyletwo}%

\theoremstyle{thmstylethree}%

\begin{document}

\title[Article Title]{Micro-Expression Recognition by Motion Feature Extraction based on Pre-training}


\author[1]{\fnm{Ruolin} \sur{Li}}\email{liruolin918@gmail.com}

\author*[1]{\fnm{Lu} \sur{Wang}}\email{wanglu@cse.neu.edu.cn}

\author[1]{\fnm{Tingting} \sur{Yang}}\email{2201900@stu.neu.edu.cn}

\author[2]{\fnm{Lisheng} \sur{Xu}}\email{xuls@bmie.neu.edu.cn}

\author[1]{\fnm{Bingyang} \sur{Ma}}\email{2101768@stu.neu.edu.cn}

\author[3]{\fnm{Yongchun} \sur{Li}}\email{liyongchun@contain.com.cn}

\author[3]{\fnm{Hongchao} \sur{Wei}}\email{dawei82@163.com}

\affil*[1]{\orgdiv{School of Computer Science and Engineering}, \orgname{Northeastern University}, \orgaddress{\city{Shenyang}, \postcode{110167}, \state{Liaoning}, \country{China}}}

\affil[2]{\orgdiv{School of Medicine and Biological Information Engineering}, \orgname{Northeastern University}, \orgaddress{\city{Shenyang}, \postcode{110167}, \state{Liaoning}, \country{China}}}

\affil[3]{\orgname{Shenyang Contain Electronic Technology Co., Ltd.}, \orgaddress{ \city{Shenyang}, \postcode{110167}, \state{Liaoning}, \country{China}}}


\abstract{Micro-expressions (MEs) are spontaneous, unconscious facial expressions that have promising applications in various fields such as psychotherapy and national security. Thus, micro-expression recognition (MER) has attracted more and more attention from researchers. Although various MER methods have emerged especially with the development of deep learning techniques, the task still faces several challenges, e.g. subtle motion and limited training data. To address these problems, we propose a novel motion extraction strategy (MoExt) for the MER task and use additional macro-expression data in the pre-training process. We primarily pretrain the feature separator and motion extractor using the contrastive loss, thus enabling them to extract representative motion features. In MoExt, shape features and texture features are first extracted separately from onset and apex frames, and then motion features related to MEs are extracted based on the shape features of both frames. To enable the model to more effectively separate features, we utilize the extracted motion features and the texture features from the onset frame to reconstruct the apex frame. Through pre-training, the module is enabled to extract inter-frame motion features of facial expressions while excluding irrelevant information. The feature separator and motion extractor are ultimately integrated into the MER network, which is then fine-tuned using the target ME data. The effectiveness of proposed method is validated on three commonly used datasets, i.e., CASME II, SMIC, SAMM, and CAS(ME)³ dataset. The results show that our method performs favorably against state-of-the-art methods.}


\maketitle

\section{Introduction}\label{sec1}
\label{}
Micro-expressions (MEs) differ from common macro-expressions in that they have some unique characteristics. First, MEs are short in duration, usually lasting only 1/25 to 1/3 seconds \cite{1}. Second, the motions of MEs are subtle and thus difficult to observe and identify. Third, MEs are generated unconsciously, which makes them difficult to conceal or disguise. Consequently, when individuals attempt to hide their emotions, their true feelings can be analyzed through MEs. Therefore, micro-expression recognition (MER) has broad applications in psychotherapy \cite{2}, national security \cite{3} and lie detection, among others \cite{4,5,6}.\par
So far, some remarkable achievements have been made for the macro-expression recognition task \cite{7,8,9,10}. However, due to the low intensity and short duration of the motion of MEs, approaches used for macro-expression recognition do not transfer well to the MER task. With the development of deep learning techniques, more and more researchers \cite{11,12,13,14,15} have proposed new MER methods according to the characteristics of MEs. Methods based on convolutional neural networks (CNNs) can be mainly divided into 2DCNN- and 3DCNN-based methods. For example, Zhou et al.\cite{11}proposed a feature refinement method that uses the 2DCNN framework to extract expression-specific features and fuse these features into the final features for expression classification. Li et al. \cite{12} utilized optical flows as network input and proposed a joint feature learning framework that combines local and global information to recognize MEs. Khor et al.\cite{13} utilized VGGNet-16 to capture the spatial features of MEs in three streams: ME sequences, optical flows, and optical strains. In terms of 3DCNN, Cai et al. \cite{14} proposed a 3D SE-DenseNet that fuses Squeeze and Excitation Network and 3D DenseNet, while adaptively weighting the features by integrating the extracted spatiotemporal information. Graph Convolutional Networks (GCN) \cite{36} has also been applied to MER tasks in recent years. Lei et al. \cite{15} proposed to represent image patches of the face by a facial graph and introduce the information of the action units into the facial graph representation by word embedding and GCN. In response to the scarcity of micro-expression datasets, Zhou et al.\cite{62} proposed a method for generating micro expression sequences using a generative adversarial network. \par
Although many MER methods have been proposed, there are still several unresolved issues associated with the MER task. First, because of the short duration, MEs can only be captured using high-speed cameras in the laboratory environment, which brings difficulty in acquiring. Meanwhile, the tiny motion intensity of MEs poses a challenge to data annotation. Both reasons result in a limited number of public data. Second, due to the difficulty of capturing subtle motion, most methods utilize optical flow or action units to extract motion features, which require additional calculations. There are few methods that can directly extract effective motion features from input frames.\par

\begin{figure}[!t]
\centering
\includegraphics[width=0.9\textwidth]{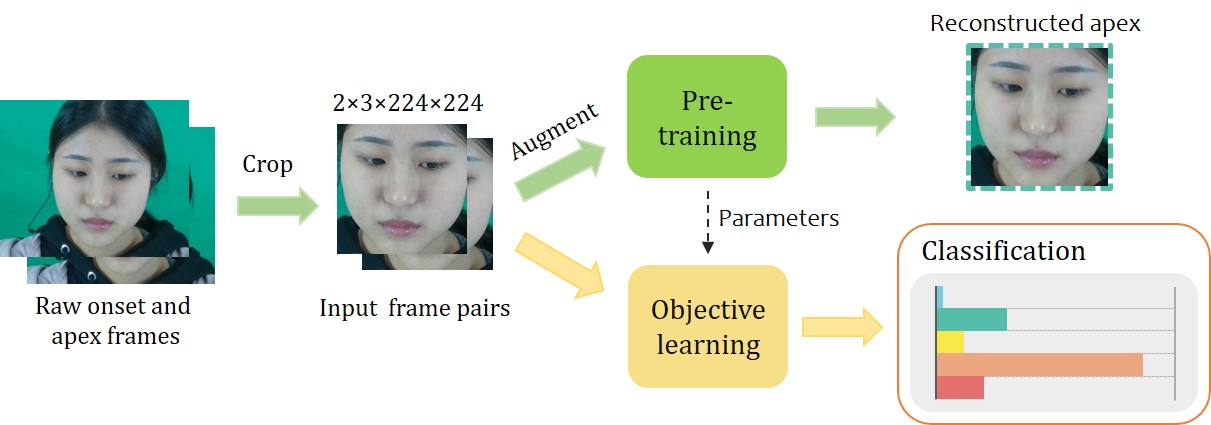}
\caption{The main steps of the proposed method: facial cropping, pre-training to reconstruct apex frames, and objective learning to classify MEs.}
\label{fig:Figure 1}
\end{figure}
In this paper, we propose a MER method that aims to addressing two of the above issues in MER, i.e., the challenge of extracting representative ME features and 
the overfitting problem caused by insufficient training data. Inspired by \cite{16} , which uses the difference between shape representations of two given frames to extract motion, we proposed a motion extraction strategy (MoExt) to extract the subtle motion between onset and apex frames for the MER task. To alleviate the overfitting problem caused by limited training data, we used the macro-expression data in the pre-training process to increase the total amount of training data.\par

Specifically, in the proposed motion extraction strategy MoExt, we first extract the shape features and facial texture features from the onset and apex frames. Then, the ME-specific motion features for MER are obtained by fusing the shape features of the onset and apex frames. Such operations can reduce the interference of the irrelevant information in the input data and thus beneficial for MER.  Since pre-training is self-supervised, a large amount of macro-expression data is added to the dataset. During the pre-training phase, we reconstructed the apex frames using motion features and facial texture features to validate the effectiveness of feature separator and motion extractor. We used the CAS(ME)³ \cite{17} dataset with a larger number of both macro- and micro-expression samples and CASME II\cite{37} as pre-training data. Finally, in the objective learning phase, the MER network that consists of shape feature extractor and a motion extractor. \par
In summary, the main contributions of this paper are as follows:
\begin{itemize}
  \item [1.]
A ME motion extraction strategy MoExt is proposed that can effectively extract the ME specific motion information from the input frames for classification. Instead of the commonly used optical flow based input \cite{11,18,32} or the video sequences \cite{49}, the input of our method is just simply the onset and apex frames.
  \item [2.]
A pre-training strategy and contrastive loss are employed to enable the MoExt to more effectively extract inter-frame subtle motion features through the reconstruction of the apex frames. Both the micro- and macro-expression data in the CAS(ME)³ dataset are utilized for the pre-training to alleviate overfitting. 
  \item [3.]
The effectiveness and robustness of the proposed MoExt based MER method are proved by comparing the results with state-of-the-art MER methods on publicly available ME datasets.
\end{itemize}\par

The organization of the paper is as follows: In section \ref{sec2}, we describe the existing works related to MER, and in section \ref{sec3}, we present in detail the proposed method. In section \ref{sec4}, we first introduce the datasets used in the experiment and provide experimental details, and then we present the experimental results. Finally, the conclusion is made in section \ref{sec5}.

\begin{figure}
\centering
\includegraphics[width=0.9\textwidth]{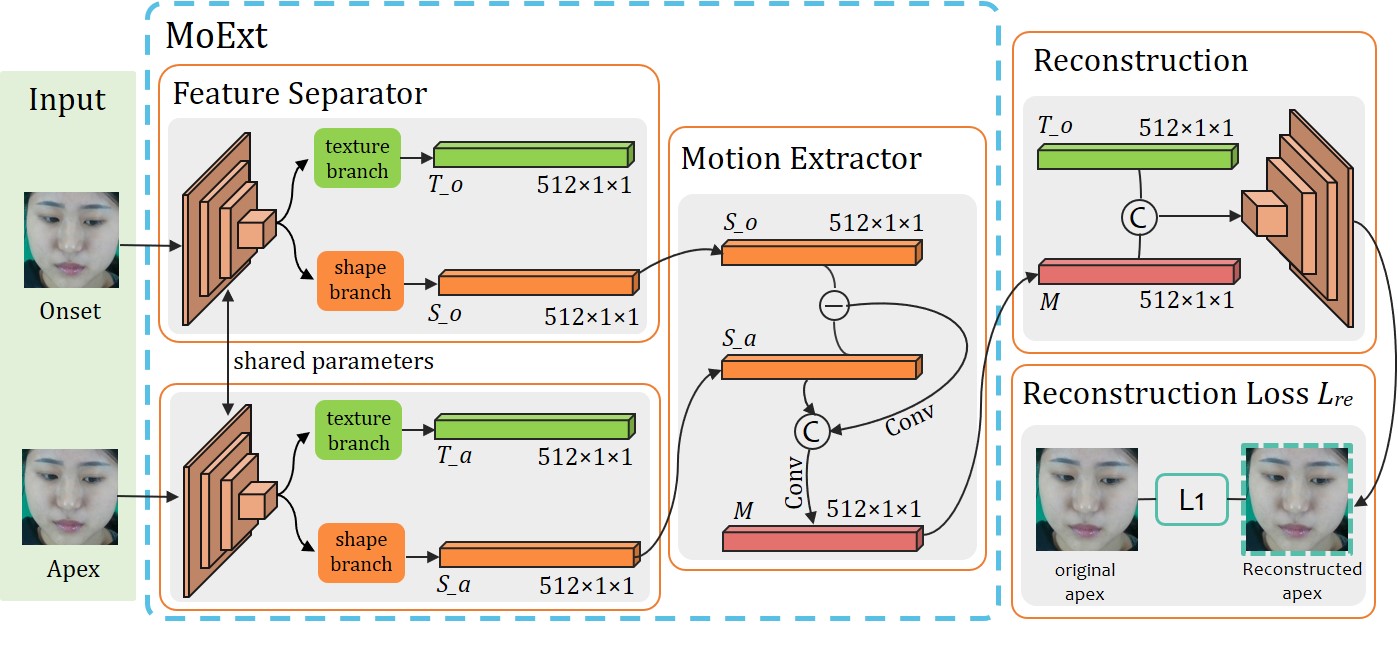}
\caption{Pre-training network framework. The feature separator is responsible for extracting shape and texture features, while the motion extractor is responsible for extracting motion features of MEs using shape features. The reconstruction module is responsible for reconstructing the apex frames. }
\label{fig:Figure 2}
\end{figure}
\section{Related work}\label{sec2}
Existing MER methods can mainly be divided into traditional machine learning and deep learning methods.\par
In traditional methods, Polikovsky et al.\cite{19} proposed a method that divides the face into specific regions and uses them to recognize MEs based on the 3D-Gradients directional histogram descriptor. Pfister et al.\cite{20} proposed the use of time interpolation model (TIM) and local binary patterns from three orthogonal planes (LBP-TOP) texture descriptor for MER. Guo et al.\cite{21}combined LBP-TOP with the nearest neighbor for MER, and obtained reasonable results. Guo et al.\cite{22} used centralized tri-orthogonal panels binary patterns (CBP-TOP) and extreme learning machine (ELM) for MER. Wang et al. proposed six-intersection LBP (LBP-SIP)\cite{23}. Compared with the LBP-TOP feature, the recognition time of \cite{23} improves by about 2.8 times, and the accuracy of LBP-TOP and LBP-SIP are almost the same. Happy et al.\cite{24} proposed to encode the temporal patterns of MEs using the Fuzzy Histogram of Optical Flow Orientations (FHOFO). The experimental results show that the performance of FHOFO is better than the best result at that time.\par
Many deep learning-based MER methods have been proposed in recent years. Among them, 2DCNN-based network structures are more common. For example, Gupta\cite{27} proposed the MERASTC framework to improve the problem of incomplete ME feature encoding and information redundancy. Zhou et al.\cite{26} proposed a feature refinement method to achieve expression-specific feature learning and fusion for MER. Li et al.\cite{12} proposed a joint feature learning framework that combines local and global information to recognize MEs using optical flow as the network input. The authors also mentioned that not all regions contribute equally to ME classification, and some regions do not even contain ME related information. Liu et al. \cite{33} took the amplitude as well as the horizontal and vertical components of optical flow as input to extract ME features using a 2DCNN network. Additionally, they employ genetic algorithms as a search-based optimization technique to improve the classification performance.\par
Since the dynamic information over time is important and affects the expressiveness of the model, each frame in the ME sequence sample has its own significance for MER. However, 2DCNN cannot obtain the spatiotemporal dynamic information simultaneously by convolution. To address this issue, some methods apply 3DCNN to the MER task. Zhao et al. \cite{32}proposed a two-stage learning method based on a siamese 3D convolutional neural network. In the first stage, the network learns common features of MEs using optical flows to determine if the input pair belong to the same category, while in the second stage, they fine-tune the structure of the model and train it with the focal loss. Zhao et al. \cite{34} proposed a MER framework that consists of a 3D residual network to learn the precise ME feature prototypes and an attention module to focus on the local facial movements.\par

In order to learn the subtle facial variations of MEs, multi-stream networks have been employed to capture the various features using different streams. Khor et al. \cite{13} used VGGNet-16\cite{53} to capture spatial features of MEs in three streams, i.e., ME sequences, optical flows and optical strain, and proposed an enriched long-term recurrent convolutional network (ELRCN) to capture the temporal information and perform the final classification. Wu et al. \cite{28} proposed a three-stream framework that combines 2D and 3DCNN (TSNN)   to capture the ME features for classification, and a temporal sampling deformation module is also proposed to normalize the time length and preserve the temporal information in ME sequences.  Shao et al. \cite{61} proposed to learn an identity-invariant representation via an adversarial training strategy, which is beneficial
for removing the interference of identity information to
MER. Song et al. \cite{29} proposed a three-stream convolutional neural network (TSCNN) consisting of static spatial stream, local spatial stream, and dynamic temporal stream, which are used to learn  facial global region, facial local region and temporal features in ME videos, respectively.\par
Above deep-learning based MER methods mainly rely on a data pre-processing stage to obtain the optical flows or facial landmarks as the network input which makes the task complex. However, inputting more frames also means inputting more facial texture information into the network, which may interfere with the feature extraction process of the network. This paper distinguishes from the existing MER methods by designing an end-to-end MER framework that uses only onset and apex frames as input, and reduces the interference of facial texture information by learning to extract effective motion information in the pre-training learning phase.
\begin{figure}[!h]
\centering
\includegraphics[width=0.9\textwidth]{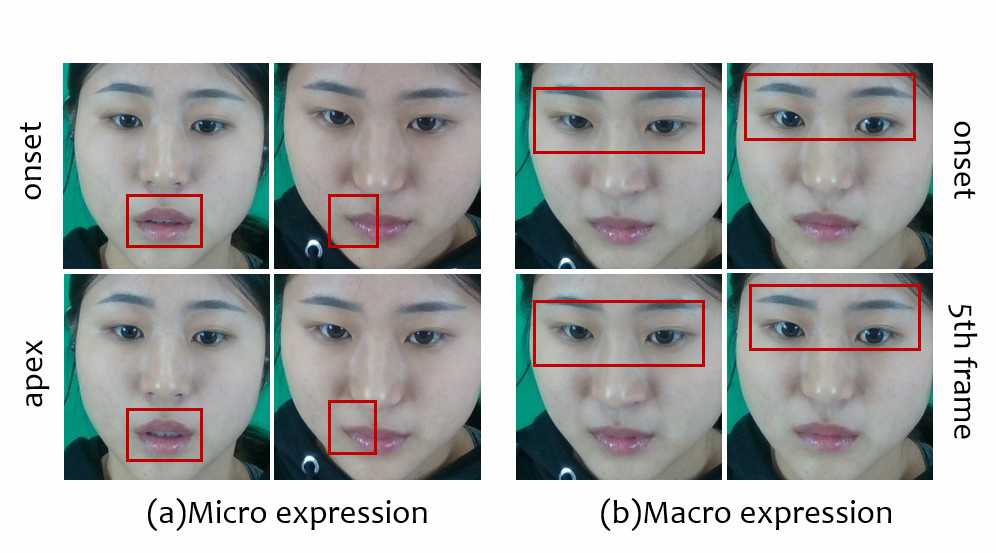}
\caption{(a) shows a comparison between the onset and apex frame of the ME, while (b) shows a comparison between the onset and fifth frame of a macro-expression. The red box marks the motion area.}
\label{fig:Figure 5}
\end{figure}
\section{Method}\label{sec3}
The main steps of the proposed MER method are shown in Figure \ref{fig:Figure 1}. Specifically, during the pre-training phase, we first utilize a large amount of facial expression data, which includes both macro- and micro-expressions, to train the feature separator and motion extractor of MoExt, as shown in Figure \ref{fig:Figure 2}. To ensure consistency in the motion ranges of the macro- and micro-expression data during training, we use the 5th frame to take the place of the apex frame as the pseudo apex frame for macro-expression data in the pre-training. As can be seen from Figure \ref{fig:Figure 5}, the 5th frame of a macro-expression has a similar motion amplitude as the apex frame of a ME.\par
The pre-training enables the network to effectively extract motion features from the onset and apex frames. Subsequently, we integrate the pre-trained feature separator and motion extractor of MoExt into the classification network, and then train it with the ME data. We will elaborate on the details of MoExt in the following subsections.\par

\subsection{Pre-training}\label{subsec3-1}
To address the issue of overfitting resulting from insufficient training data, we employ both micro- and macro-expression samples from an extra dataset as the training data during the pre-training phase. However, since the motion range of the macro-expression apex frames are usually very high, we use the N-th frame ( using the 5th frame in the experiment ) of each macro-expression sample to take the place of the apex frame (hereafter referred to as the apex frame).\par

Since MEs have very subtle motion intensity that is difficult to capture, it is crucial to equip the network with the ability to extract the motion features specific to MEs while discarding irrelevant information from the input for further classification. In order to effectively separate the motion and facial texture features of the ME, we use the onset and apex frames as input in the pre-training stage, and reconstruct the apex frames using the extracted motion features and texture features, as shown in Figure \ref{fig:Figure 2}.\par

In particular, first, we utilize the feature separator to extract the shape features and texture features from the onset and apex frames. The shape features of the input frames are subsequently utilized to extract motion features and reconstruct the apex frames, while the texture features of the onset frame are used for the reconstruction of the apex frames. The effective extraction of shape features and facial texture features is achieved by enforcing that the reconstructed apex frame is similar to the apex frame at the pixel level.\par
The apex frame reconstructed in pre-training is defined as :
\begin{equation}
\widetilde{X}{\_a}=R\left(E\left(F\left(X\_o\right),F\left(X\_a\right)\right),F\left(X\_o\right)\right)\label{eq1}
\end{equation}
where $F$ represents the feature separator, $E$ denotes the motion extractor, $R$ is the apex frame reconstruction module,\textbf{ \textit{$X\_o$}} and\textbf{ \textit{$X\_a$}} are the input onset and apex frames, and \textbf{ \textit{$\widetilde{X}{\_a}$}} is the reconstructed apex frame. In the following subsections, we will introduce the feature separator $F$, motion extractor $E$  and apex frame reconstruction module  $R$  in detail.\par

\begin{figure}[!t]
\centering
\includegraphics[scale=0.6]{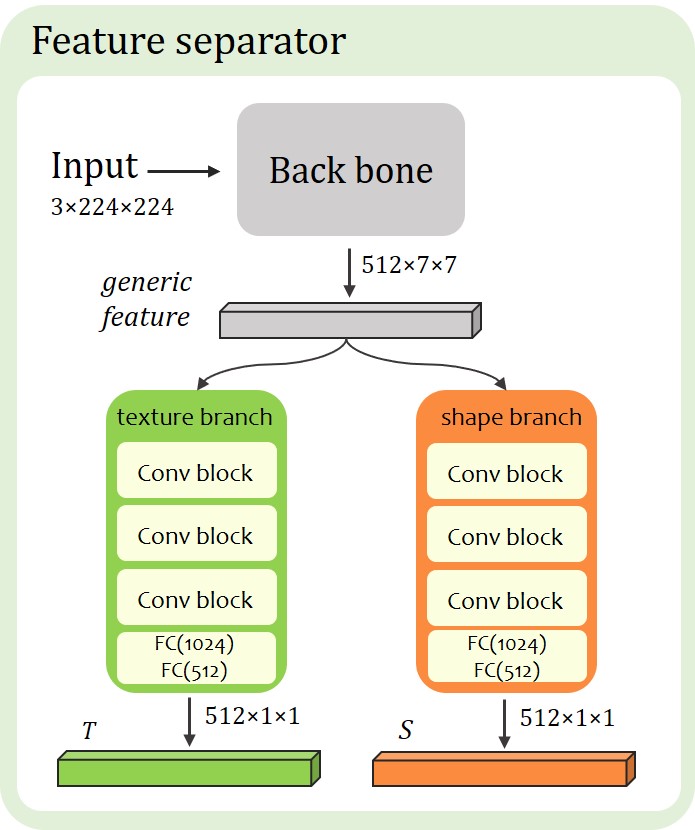}
\caption{Overview of the feature separator structure. The feature separator is responsible for separating the shape features and texture features. In this case, the generic features are extracted by the backbone, and then fed to the shape branch and texture branch to extract shape features and texture features, respectively. }
\label{fig:Figure 4}
\end{figure}

\subsection{Motion extraction strategy MoExt}\label{subsec3-2}
During the pre-training phase, MoExt mainly consists of a feature separator and a motion extractor. In the objective learning phase, the feature separator transforms into a shape feature extractor by removing the texture branch and inherits the parameters from the pre-training phase.

\subsubsection{Feature separator}\label{subsubsec3-2-1}
We design a feature separator, as shown in Figure \ref{fig:Figure 4}. The role of the feature separator is to extract and separate texture features and shape features from input frames. It consists of three parts, including a backbone network for extracting generic features, as well as texture and shape branches. generic features are input into the texture and shape branches, and after multiple convolution operations, texture features and shape features are obtained. These two branch structures are identical, both composed of Conv blocks and fully connected layers. Each Conv block includes two 2D convolutions, two batch normalizations, and ReLU activation functions, with the specific structure shown in the Table \ref{tab:Table 1}.\par

\subsubsection{Motion extractor}\label{subsubsec3-2-2}
Inspired by Tae et al. \cite{16}, we design a motion extractor module for MER, as shown in Figure \ref{fig:Figure 2}. The structure of each Conv block is specified in Table \ref{tab:Table 1}. Considering that the apex frame contains the highest motion intensity, while the onset frame is a neutral expression that can be used as a reference, we extract ME related motion features by utilizing the shape features of both the onset and apex frames.\par
Specifically, the motion extractor takes the shape features $S\_o$ from the onset frame and the shape features $S\_a$ from the apex frame as inputs, and subtracts $S\_a$ from $S\_o$ element-wise. Apart from the the absolute value of the subtracted features, the shape feature of apex frame also contains crucial semantic information regarding MER and apex reconstruction. Thus, we fuse these pieces of information by concatenating along the channel dimension. Retaining the shape features of the apex frame enables us to achieve better reconstruction of apex frame. The concatenated features are fed into Conv block to obtain the final motion features $M$. The extracted motion features $M$ can be defined as :
\begin{equation}
\Delta S=\left|S\_a-S\_o\right|\label{eq2}
\end{equation}
\begin{equation}
M=C_2\left(Concat\left(C_1\left(\Delta S\right),S\_a\right)\right)\label{eq3}
\end{equation}
where \textbf{ \textit{\textbf{$S\_a$}}} and \textbf{\textit{\textbf{$S\_o$}} }are the shape features extracted from apex and onset frames, respectively, \textbf{ \textit{\textbf{$C_{i}$}}} denotes the $i$-th Conv block.\par
\subsection{Apex reconstruction}\label{subsec3-3}
In order to ensure the motion and texture features are properly separated, we propose a module that utilizes the texture features  $T\_o$ obtained from the onset frame and the motion features $M$ obtained from the motion extractor to reconstruct the apex frame. By constraining the reconstructed apex frames similar to the apex frames at pixel level, the motion features can be effectively separated from the texture features. The structure of the apex frame reconstruction module is shown in Table \ref{tab:Table 2}. \par
The reconstruction module consists of several Conv blocks, and the final output of the reconstructed apex frame module $\widetilde{X}\_{a}$ is defined as :
\begin{equation}
\widetilde{X}\_{a}=R\left(Concat\left(M,T\_o\right)\right)\label{eq4}
\end{equation}
where  \textit{\textbf{$T\_o$}}  is the texture feature extracted by the feature separator from the onset frame.

\begin{figure}[!t]
\centering
\includegraphics[width=0.9\textwidth]{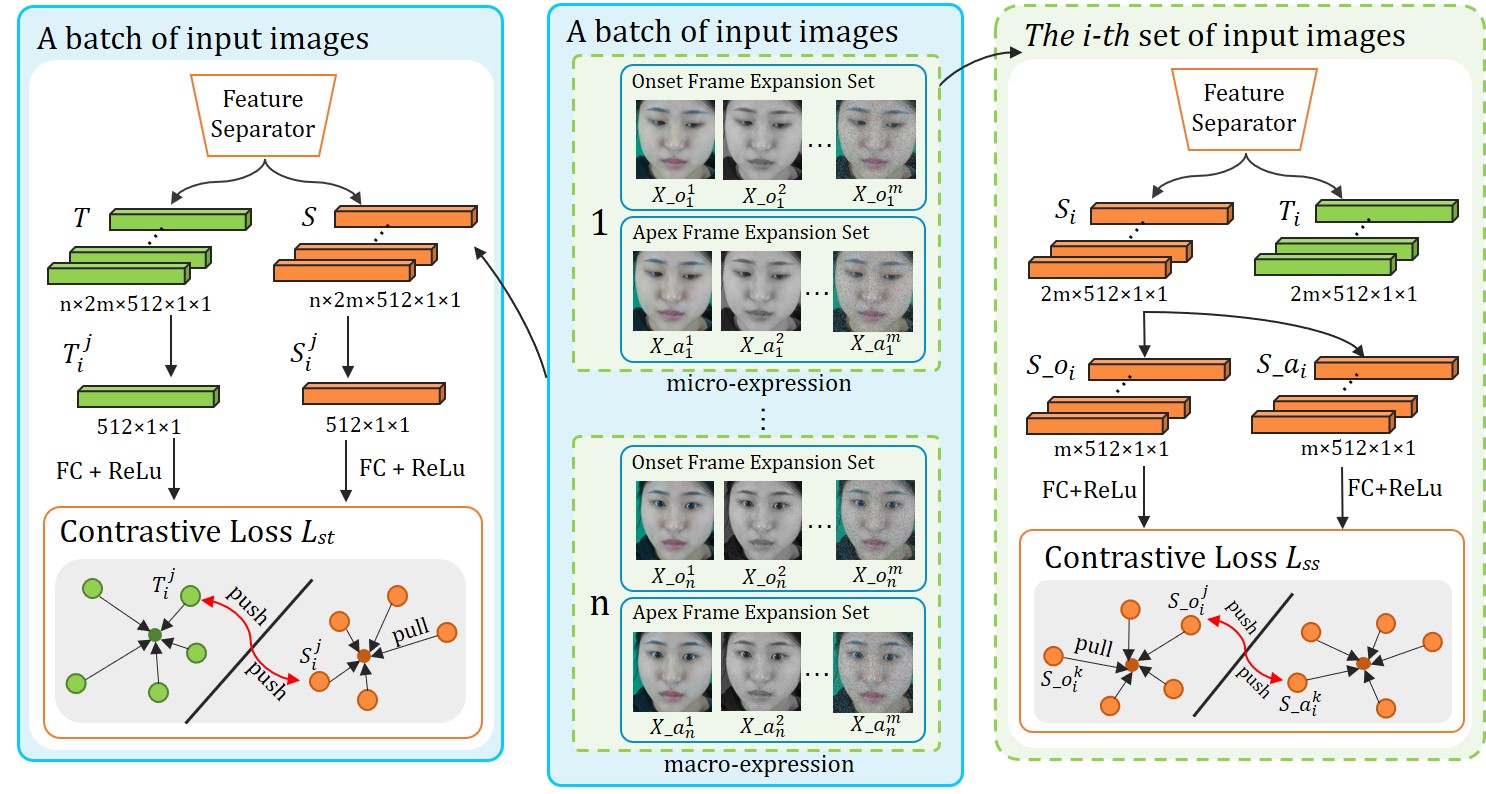}
\caption{Illustration of contrastive losses, with the middle section providing an explanation of the inputs, the left side illustrating the contrastive loss \( L_{\text{st}} \) for texture and shape features, and the right side illustrating the contrastive loss \( L_{\text{ss}} \) for shape features between onset and apex frames.}
\label{fig:Figure 3}
\end{figure}

\subsection{Objective learning}\label{subsec3-4}
With the pre-training process in MoExt described above, the module has already been equipped with the ability to extract motion features between two frames, and hence can be utilized for MER. Based on pre-training network structure, we propose a ME classification network.\par
In particular, the texture branch is removed from the feature separator, while the motion extractor maintains the same structure as in the pre-training phase. Unlike the pre-training phase, the training data only include the mirco-expression data. Finally, the extracted motion features are utilized for MER.
\subsection{Loss function}\label{subsec3-5}
In this paper, we use the commonly used cross-entropy loss as the loss function during the objective learning phase.
In the pre-training phase, since we want to obtain the motion features based on the shape information extracted from the onset and apex frames, while avoiding the influence of irrelevant texture information, the following loss functions are used.  Firstly, as the reconstructed apex frame $\widetilde{X}\_{a}_{i}$ should resemble the apex frame $X\_{a}_{i}$, the loss function for each batch is defined as:
\begin{equation}
L_{re}=\frac{1}{n}\cdot \sum_{i=1}^{n} \left|X\_{a}_{i}-\widetilde{X}\_{a}_{i}\right|\label{eq5}
\end{equation}
where $i \in\ \{1,2,3,...,n\}$ denotes the index within a batch.
\par

To enhance the effectiveness and robustness of the feature separator, we augment the input data, perform feature separation on the augmented data, and calculate contrastive losses, as shown in Figure \ref{fig:Figure 3}. Specifically, we apply \textit{m-1} types of data augmentation operations to the pairs of onset and apex frames, resulting in corresponding Onset Frame Expansion Sets $X\_o_i$ and Apex Frame Expansion Sets $X\_a_i$, forming the $i$-th set of inputs. Each expansion set contains \textit{m} instances, thus the $i$-th set of inputs comprises \textit{2m} instances. We define two contrastive losses: \( L_{\text{st}} \) is the contrastive loss for shape and texture features, and $L_{ss}$ calculates the contrastive loss for the shape features between the onset frame expansion set and the apex frame expansion set.\par

Firstly, to ensure the texture and shape features are well separated, we map the shape and texture features from the same instance ( such as $S\_o_i^j,T\_o_i^j$ ) into the high-dimensional space and consider shape and texture features as positive and negative samples, respectively. By minimizing the similarity between the texture and shape features in the high-dimensional space, we expect to make the shape features of the same instance contain less facial texture information, thus reducing the interference of redundant information in the final classification. 
\par

We define the loss function \( L_{\text{st}} \)  to measure the contrastive loss between shape features and texture features, where  \(\bar{S}\)  represents the anchor point of the contrastive loss, obtained by computing the average of shape features in a batch. By minimizing the distance between shape features and the anchor point, while magnifying the distance between texture features and the anchor point, we encourage better decoupling of shape and texture features. Each input set comprises an onset frame expansion set and an apex frame expansion set, and each expansion set contains \textit{m} instances. Thus, the $i$-th set of input images comprises \textit{2m} instances, corresponding to \textit{2m} pair of shape-texture features. \( L_{\text{st}} \) is formally defined as:
\begin{equation}
\begin{split}
L_{st}=\frac{1}{n}\cdot \frac{1}{2m}\cdot \sum_{i}^{n}\sum_{j}^{2m}{max(0,||{f(\bar{S})-f(S_i^j)||}_2-||{f(S_i^j)-f(T_i^j)||}_2+\varepsilon)}\label{eq6}
\end{split}
\end{equation}

\begin{equation}
\bar{S}=\frac{1}{n}\cdot\frac{1}{2m}\cdot\sum_{i}^{n}\sum_{j}^{2m}{S_i^j}\label{eq7}
\end{equation}
where $S_i^{j}$ and $T_i^{j}$ denote the shape feature and the texture feature extracted from the $i$-th set of input images, respectively; $f(\cdot)$ represents the function that maps all features into the high-dimensional space; $m$ is the size of the expansion set, and $\epsilon\ \in\ [0.2, 0.5]$ represents the minimum distance difference.\par

Meanwhile, to better extract the shape features, we map the extracted shape features into a high-dimensional space and define the loss function $L_{ss}$ between the onset and apex frames of the same subject. For a set of inputs, when the shape features come from the same expansion set, they should be as similar as possible in high-dimensional space since the instances within the expansion set are all augmented examples of the same frame. On the other hand, when the shape features come from the different expansion set, they should be as dissimilar as possible. We define the contrastive loss $L_{ss}$ as follows:\par

\begin{equation}
\begin{split}
L_{ss} = \frac{1}{n} \cdot \frac{1}{2m} \cdot \frac{1}{2m} \cdot \sum_{i}^{n} \sum_{j}^{2m} \sum_{k}^{2m} \left( \left[y_j=y_k\right]||f(S_i^j)-f(S_i^k)||_2 \right. \\
+ \left. \left[y_j\neq y_k\right]\max(0, \varepsilon - ||f(S_i^j)-f(S_i^k)||_2) \right)
\end{split}
\end{equation}
where $y_j, y_k \in\ \{onset, apex\}$ denote the frame set corresponding to the instance.\par

\par
In summary, the total loss function in the pre-training phase is defined as:
\begin{equation}
L=L_{re}+\alpha_1\times L_{st}+\alpha_2\times L_{ss}\label{eq9}
\end{equation}
where $\alpha_i\in[0,1]$ is the corresponding loss function weight value.
\section{Experiments}\label{sec4}
In this section, we first describe the experimental setup, including the datasets, data pre-processing, parameter settings, and evaluation metrics. Then we will compare the results of the proposed MoExt with those of the state-of-the-art MER methods. Finally, we will present the results of the ablation studies.\par
\subsection{Experiments setting}\label{subsec4-1}
\subsubsection{Datasets}\label{subsubsec4-1-1}
We conducted experiments on four datasets: CASME II, CAS(ME)³, SAMM\cite{38}, and SMIC-HS\cite{39}. These datasets will be described in detail below.\par
The SMIC (Spontaneous Micro-expression Corpus) dataset is released in 2013 and contain three sub-datasets, namely HS, VIS, and NIR, which differ by the type of camera used. SMIC-HS uses a high frame rate camera and contains 164 ME samples. The data is categorized into three emotion types: positive, negative, and surprised, and is captured at a resolution of 640×480 and a frame rate of 25 frames per second.\par
CASME II is released in 2014, and the data are captured using a high-speed camera with a frame rate of 200 frames per second. The dataset comprises 26 subjects and 247 samples, with an image resolution of 640×480. The MEs are divided into 7 emotional types: Happiness (33), Repression (27), Surprise (25), Disgust (60), Fear (2), Sadness (7), and Others (102). In addition to the labels for the emotion types, labels of the action units (AUs), the onset, apex, and offset frames for each sample are also provided by the dataset.\par
SAMM (Spontaneous Actions and Micro-Movements) dataset is released in 2016, consisting of 159 ME samples and 32 subjects, with a sample resolution of 2040×1088 and 200 frames per second. The subjects in this dataset come from 13 races. The sample labels are comprised of 8 emotional types: Happiness (26), Fear (8), Surprise (15), Anger (57), Disgust (9), Sadness (6), Contempt (12), and Others (26). Additionally, the dataset provides onset, apex and offset frame serial numbers of each sample, as well as AU labels.\par
CAS(ME)³ is a large dataset released in 2022, divided into three parts, A,B, and C, with 247 subjects, 112 males and 135 females, all Asian. Part A consists of 943 ME samples and 3143 macro-expression samples, with 100 subjects' micro-expression samples labeled with 7 emotion types: Happiness (64), Disgust (281), Fear (93), Anger (70), Sadness (64), Surprise (201), and Others (170). Part B includes 116 subjects and 1508 unlabeled video clips. Part C consists of 116 ME samples and 347 macro-expression samples. Both A and B are expressions induced by using emotion-based videos, while C comprises MEs induced by simulating crime patterns. The samples are captured at a resolution of 1280×720 and a frame rate of 30 frames per second. Along with the sample categories, onset, apex, and offset frame serial numbers of each sample, the dataset also provides AU labels, depth information, and other physiological information. In the experiments, we utilizes Part A of the dataset.\par
\begin{table}[h]
\caption{Structure of the Conv block, where \textit{out} represents the number of output channels}\label{tab:Table 1}%
\begin{tabular}{@{}ll@{}}
\toprule
Architecture of the Conv block & Output\\
\midrule
Conv(\textit{out,3,1}), BN, ReLU & \textit{out×h×w}\\
Conv(\textit{out,3,1}), BN, ReLU & \textit{out×h×w} \\
\botrule
\end{tabular}
\end{table}

\begin{table}[t]
\caption{Architecture of the reconstruction module. The parameter of the Conv block denotes the number of output channels and the two parameters of Avg pooling denote kernel size and stride. The three parameters of Conv2d correspond to the number of output channels, kernel size, and stride.}\label{tab:Table 2}%
\begin{tabular*}{\textwidth}{@{\extracolsep\fill}lc}
\toprule%
Architecture of reconstruction module & Output\\
\midrule
Concat & $1024\times 1\times 1$\\
$\left\{ \text{Up sampling(2), Conv2d(512,3,1), BN, ReLU} \right\} \times  3$
&  $512\times 8 \times 8$ \\
Avg pooling(2,1) & $512\times 7\times 7$ \\
Up sampling(2) Conv2d(512,3,1),BN,ReLU & $512\times 14\times 14$ \\
Up sampling(2) Conv2d(256,3,1),BN,ReLU & $256\times 28\times 28$ \\
Up sampling(2) Conv2d(128,3,1),BN,ReLU & $128\times 56\times 56$ \\
Up sampling(2) Conv2d(64,3,1),BN,ReLU & $64\times 112\times 112$ \\
Conv block(32)×5 & $32\times 112\times 112$ \\
Up sampling(2) & $32\times 224\times 224$ \\
Conv2d(512,3,1), ReLU & $3\times 224\times 224$ \\
\botrule
\end{tabular*}
\end{table}
\subsubsection{Experimental settings}\label{subsubsec4-1-2}
Both micro- and macro-expressions are included in the experiments, with the data in CASME II and CAS(ME)³-A being used in the pre-training stage.
We employ the Optical flow (OF) to quantify the motion amplitude of micro-and macro-expression. OF is a reliable approximation method for estimating the motion of two-dimensional images. By utilizing the temporal variations of pixels in an image sequence and the correlation between adjacent frames, it establishes correspondences between the previous and current frame to calculate the motion information between adjacent frames. To visually observe the magnitude and angle of motion, we convert $u$ and $v$ of the OF into polar coordinates.\par
\begin{figure}[!t]
\centering
\includegraphics[width=0.9\textwidth]{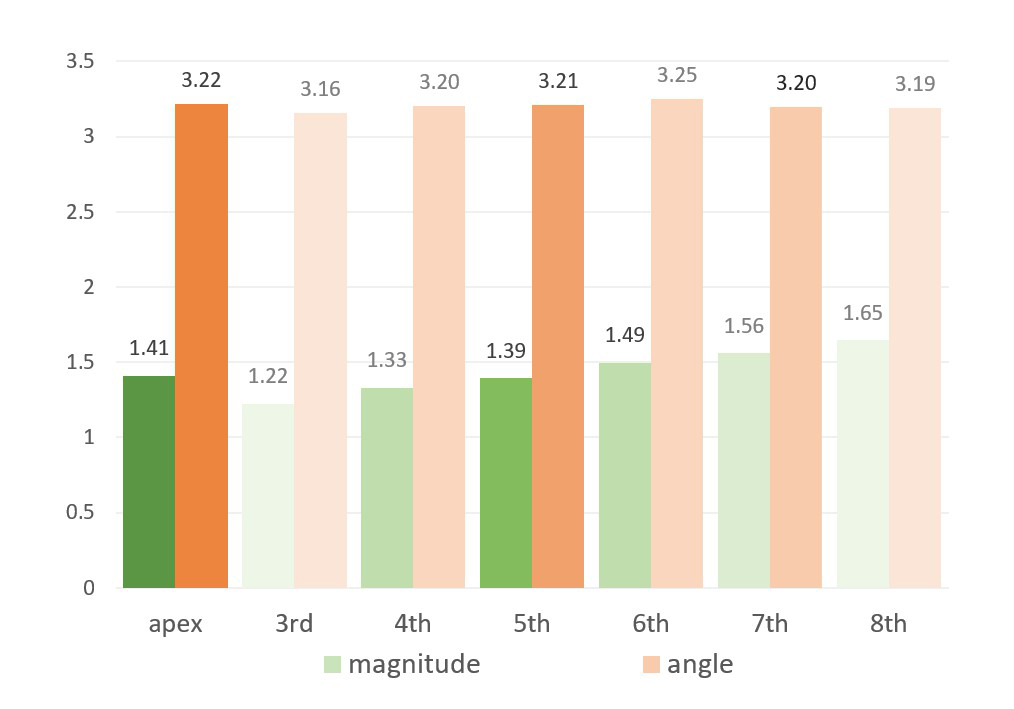}
\caption{In the figure, the green portion represents the average magnitude of the optical flow calculated between the 1st and each frame, while the orange portion represents the average angle. The apex column corresponds to the optical flow between the onset and apex frame of the ME.}
\label{fig:macro}
\end{figure}

 
In Figure \ref{fig:macro}, as the color intensity gets closer to the apex column, the change intensity of the frame also gets closer to the apex frame. Consequently, it can be inferred that frame 5 exhibits the highest similarity in terms of both amplitude and angle with the apex frame of ME. Hence, in the pre-training stage, the 5th frames are chosen to substitute the apex frames of the macro-expressions to expand the training data.\par
Since SMIC does not provide apex frame annotations, the middle frame of the sequence is used as the apex frame in our experiments. The Dlib algorithm \cite{48} is used for face localization, with the 8th, 9th, 25th, 40th, and 43rd facial landmarks utilized for alignment and cropping. To augment the data, the training data is mirrored and rotated [-10°, 10°], resulting in ten times the original data size.\par
For the pre-training phase and objective learning stage, the batch size and epoch are set to 20 and 30, respectively. The optimizer chosen for both stages is ADAM with the learning rate of 0.0001, weight decay of 0.0001, and the momentum of 0.9. The size of the input image is set to 224 × 224. The data augmentation operations for obtaining the View Expansion Set include adding random noise, increasing image contrast, and converting images to black and white. In the experiments, two operations are randomly selected, that is, m = 3. The weight $\alpha_i $ (i=1,2) of the loss function are set to 0.5 and 1 respectively. The Convolution operation is represented by Conv, and the structure of the Conv block structure is provided in Table \ref{tab:Table 1}, where the required parameters for the operation are labeled. The structures of the apex frame
reconstruction module in the pre-training stage is shown in Table \ref{tab:Table 2} , respectively. The experiments are conducted using PyTorch on Linux 18.04, with the utilization of the NVIDIA GeForce GTX 3090 24 GB GPU.\par

\subsubsection{Evaluation metrics}\label{subsubsec4-1-3}
The experiments are evaluated using the leave-one-subject-out (LOSO) method, where the samples of one subject are used as the test set while the samples of the remaining subjects are used as the training set. We employ two protocols, Composite Database Evaluation (CDE) and Sole Database Evaluation(SDE). For CDE, a total of 68 subject sample sets from three datasets (SMIC-HS, CASME II, and SAMM) are mixed for training and evaluation. For SDE, the training and evaluation are carried out within one dataset.\par
Regarding the evaluation metrics, we adopt the unweighted F1 score (UF1) and the unweighted average recall (UAR), which are employed by MEGC 2019\cite{50} to assess the model performance. Furthermore, we also utilize accuracy (ACC) as an evaluation metric. The calculations of these evaluation metrics are as follows:\par
\begin{equation}
UAR=\frac{1}{C}\sum_{i=1}^{c}\frac{{TP}_i}{{TP}_i+{FN}_i}\label{eq10}
\end{equation}
\begin{equation}
F1=\ \frac{2TP}{2TP+FP+FN}\label{eq11}
\end{equation}
\begin{equation}
UF1=\frac{1}{C}\sum_{i=1}^{c}{F1}_i\label{eq12}
\end{equation}
\begin{equation}
ACC=\frac{TP+TN}{N}\label{eq13}
\end{equation}
where \textit{TP} represents true positive samples,\textit{ FP} represents the false positive samples, \textit{TN} represents the true negative samples, \textit{FN} represents the false negative samples, and \textit{C} is the number of categories classified.
\subsection{Experimental results}\label{subsec4-2}
In this subsection, we compare the results of the proposed MoExt with those of state-of-the-art methods on the CASME II, SAMM, and CAS(ME)³ datasets.  According to the common practice, the SDE evaluation strategy and the evaluation metrics ACC and UF1 are used. In addition, the CDE strategy and the UAR and UF1 metrics are applied on the composite dataset.\par
\subsubsection{Sole Database Evaluation (SDE)}\label{subsubsec4-2-1}
The results of state-of-the-art MER methods on the five emotion classifications task of the CASME II dataset are presented in Table \ref{tab:Table 4}. It can be seen that the ACC and UF1 of our method are 0.8179 and 0.7825, respectively, with UF1 outperforming all the other methods with a large margin. Among them, the FRL-DGT method also employs reconstruction methods during the pre-training phase. However, unlike MoExt, 
 FRL-DGT aims to extract displacement feature maps as its pre-training objective. The results demonstrate the effectiveness of our proposed MoExt.\par
 \begin{table}[h]
\caption{Experiment on CASME II with 5 classes}\label{tab:Table 4}%
\begin{tabular}{@{}lll@{}}
\toprule
Methods&ACC&UF1\\
\midrule
GEME(2020) \cite{41}&0.7520&0.7354\\
MiMaNet(2021) \cite{42}&0.7990&0.7590\\
LR-GACNN(2021) \cite{43}&0.8130&0.7090\\
AMAN(2022) \cite{44}&0.7540&0.7100\\
FRL-DGT(2022) \cite{56}&0.7570&0.7480\\
GLEFFN(2023) \cite{58}&0.7607&0.7564\\
C3DBed(2023) \cite{59}&0.7764&0.7520\\
MoExt(ours)&\textbf{0.8179}& \textbf{0.7825} \\
\botrule
\end{tabular}
\end{table}

 \begin{table}[h]
\caption{Experiment on SAMM with 5 classes}\label{tab:Table 5}%
\begin{tabular}{@{}lll@{}}
\toprule
Methods&ACC&UF1\\
\midrule
 MTMNet (2020) \cite{46}&0.7410&0.7360\\
 MiMaNet(2021) \cite{42}&0.7670&0.7640\\
 GARPH-AU (2021) \cite{15} &0.7426&0.7045\\
 AMAN(2022) \cite{44}&0.6885&0.6700\\
 C3DBed(2023) \cite{59}&0.7573&0.7216\\
 CMNet(2023) \cite{60}&0.7868&0.7719\\
 MoExt(ours)& \textbf{0.7903} &\textbf{0.7875}\\
\botrule
\end{tabular}
\end{table}

\begin{table}[h]
\caption{Experiment on CAS(ME)³ with 3 classes}\label{tab:Table 6}%
\begin{tabular}{@{}lll@{}}
\toprule
Methods&UF1&UAR\\
\midrule
STSTNet(2019) \cite{47}&0.3795&0.3792\\
RCN-A(2020) \cite{54}&0.3928&0.3893\\
FeatRef(2022) \cite{11}&0.3493&0.3413\\
HTNet(2023) \cite{63}&\textbf{0.5767}&0.5415\\
MoExt(ours)&0.5457& \textbf{0.5784} \\
\botrule
\end{tabular}
\end{table}

The results of the five emotion classifications conducted on the SAMM dataset are presented in Table \ref{tab:Table 5}. Our method outperforms the compared methods in terms of ACC and UF1.
The MTMNet method also takes the onset and apex frame as inputs. It decomposes the features into facial expression-related features and neutral features, leveraging the MacroNet to guide the network in learning facial expression-related features. Our method performs better, indicating that the proposed MoExt can more effectively extract motion features related to MEs.\par

Table  \ref{tab:Table 6} shows the experimental results of our method on CAS(ME)³, where the UF1 and UAR scores reached 0.5457 and 0.5784, respectively. The UAR score is significantly higher than those of the other methods listed, demonstrating the effectiveness of our MoExt method. Our UF1 score is lower than that of HTNet. We believe this is because HTNet focuses more on local motion information, and since micro-expressions are primarily localized movements.

\subsubsection{Composite Database Evaluation(CDE)}\label{subsubsec4-2-2}
Composite dataset evaluation is another widely used evaluation strategy in MER tasks. In accordance with the MEGC2019 criteria, we combine all samples from three datasets, namely SMIC-HS, CASME II, and SAMM, to create a composite dataset. We then compare the proposed MoExt with the state-of-the-art methods over the last three years on the composite dataset.\par
The experimental results of the compared methods and the proposed MoExt for the 3-class MER task on the composite dataset are presented in Table \ref{tab:Table 7}. The UF1 and UAR of MoExt achieve 0.8149 and 0.8115, both surpassing the compared methods. Among these methods, MERSiamC3D, FeatRef, and IncepTR extract optical flow before using the neural network for classification.  RCN-A, ME-PLAN and GLEFFN use ME sequences as input. The results show that the motion features we extract from onset and apex frames are relatively superior to those extracted from the full ME sequences. Similar to our method, GARPH-AU also uses the onset and apex frames as the input. It focuses on the facial action units, which demonstrate the effectiveness of our MoExt. C3DBed employs the apex frame as input and utilizes a transformer to encode facial patches. In contrast to our method, it lacks the onset frame as a neutral reference.\par

\begin{sidewaystable}

\caption{Experiment on CDE with 3 classes}\label{tab:Table 7}
\begin{tabular*}{\textwidth}{@{\extracolsep\fill}lcccccccc}
\toprule%
& \multicolumn{2}{c}{3DB-combined} & \multicolumn{2}{c}{SMIC-HS} & \multicolumn{2}{c}{CASME II} & \multicolumn{2}{c}{SAMM} \\\cmidrule{2-9}%
Methods &UF1&UAR&UF1&UAR&UF1&UAR&UF1&UAR \\
\midrule
RCN-A(2020) \cite{54}&0.7432&0.7190&0.6326&0.6441&0.8512&0.8123&0.7601&0.6715\\

MERSiamC3D(2020) \cite{32}&0.8068&0.7986&0.7356&0.7598&0.8818&0.8763&0.7475&0.7280\\

GARPH-AU (2021) \cite{15}&0.7914&0.7933&0.7292&0.7215&0.8798&0.8710&0.7751&0.7890\\

FeatRef(2022) \cite{11}&0.7828&0.7832&0.7011&0.7083&0.8915&0.8873&0.7372&0.7155\\
           
ME-PLAN(2022) \cite{34}&0.7878&0.8041&N/A&N/A&0.8941&0.8962&0.7358&0.7687\\

FRL-DGT(2022) \cite{56}&0.8120&0.8110&0.7430&0.7490&\textbf{0.9190}&0.9030&0.7720&0.7580\\

IncepTR(2023) \cite{57}&0.7530&0.7460&0.6550&0.6500&0.9110&0.8960&0.6910&0.6940\\

GLEFFN(2023) \cite{58}&0.8121&0.8208&0.7714&0.7856&0.8825&\textbf{0.9110}&0.7458&0.7843\\

C3DBed(2023) \cite{59}&0.8075&0.8013&0.7760&0.7703&0.8978&0.8882&0.8126&0.8067\\

MoExt(ours)&\textbf{0.8301}&\textbf{0.8226}&\textbf{0.7976}&\textbf{0.7803}&0.8992&0.8912&\textbf{0.8135}&\textbf{0.8100} \\
\botrule
\end{tabular*}
\end{sidewaystable}

It can also be seen that our method achieves the best result on the SAMM dataset but not on the CASME II dataset. The reason we think is as follows. The SAMM dataset consists of samples from diverse races and age groups, which demands higher generalization capability of the model. Our proposed MoExt aims at extracting general ME motion features by also incorporating the macro-expression data for pre-training and thus equip higher generalization capability. The results on the more difficult SAMM dataset demonstrate the effectiveness of our method.
\subsection{Ablation experiments}\label{subsec4-3}
In this subsection, we conduct ablation experiments on the composite dataset to demonstrate the effects of the pre-training, the motion extractor, the macro-expression data, and the contrastive loss $L_{st}$ and $L_{oa}$.\par
\begin{table}[h]
\caption{Ablation experiments on the composite dataset}\label{tab:Table 8}%
\begin{tabular}{@{}llll@{}}
\toprule
Methods&ACC&UF1&UAR\\
\midrule
w/o pre-training&$0.7812$&$0.7746$&$0.7776$\\
w/o macro-expressions&0.8023&0.8164&0.8020\\
w/o motion extractor&0.8177&0.8082&0.8038\\
w/o $Loss_{st}$ & $0.8233$ & $0.8017$& $0.8045$ \\
w/o $Loss_{oa}$ &0.8250&0.8111&0.8010\\
MoExt&\textbf{0.8417}&\textbf{0.8301}&\textbf{0.8226} \\
\botrule
\end{tabular}
\end{table}

As shown in Table \ref{tab:Table 8}, compared to the experimental results without pre-training, the results with pre-training are 0.0605, 0.0555, and 0.0450 higher in terms of ACC, UF1, and UAR, respectively, indicating the effectiveness of the pre-training. Simultaneously, we attempted to utilize only ME samples for two-stage training. The experimental results demonstrate that incorporating macro expressions into the experiment expands the training data and effectively improves the problem of overfitting. To show the necessity of the motion extractor, we also use the absolute difference between the shape features extracted from the onset and apex frames of the same sample to replace it. From the experimental results, we see that, by using the motion extractor, the three metrics can be improved by 0.0240, 0.0219, and 0.0188, respectively, which confirms the effectiveness of the motion extractor. Regarding contrastive losses $L_{st}$ and $L_{oa}$, their purpose is to enforce constraint on the learning of the motion extractor. From the perspective of evaluation metrics, the use of contrastive losses has shown a significant improvement in experimental results.\par
\section{Conclusion}\label{sec5}
This paper presents a novel motion extraction method MoExt based on pre-training for extracting subtle motion features of MEs from the onset and apex frames. The MoExt consists of feature separators and a motion extractor. The feature separator is utilized to separate the shape features and texture features, and a motion extractor is designed to extract motion features from shape features. To ensure that MoExt effectively separates shape and texture features and extracts motion features, we employ a pre-training strategy. This strategy utilizes motion features and texture features to reconstruct the apex frame and is trained using contrastive loss. In the objective learning stage, the pre-trained MoExt is
finally embedded into the MER network and fine-tuned. We evaluate our proposed MoExt using several public ME datasets. The results demonstrate its effectiveness and superiority over existing MER approaches.

\section*{Declarations}

\textbf{Funding} This work was supported by the Shenyang Science and Technology Plan Fund (No. 21-104-1-24), the National Natural Science Foundation of China (No. U21A20487 and No. 62273082), the Natural Science Foundation of Liaoning Province (No. 2021-YGJC-14), the Basic Scientific Research Project (Key Project) of Liaoning Provincial Department of Education (LJKZ00042021), and Fundamental Research Funds for the Central Universities (No. N2119008).\\
\textbf{Conflict of interest} The authors declare no conflict of interest.\\
\textbf{Ethics approval and consent to participate} Not applicable.\\
\textbf{Consent for publication} The manuscript has not been sent to any other journal for publication.\\
\textbf{Data availability} The data that support the findings of this study are available from CASME II\cite{37}, SAMM\cite{38}, SMIC\cite{39} and CAS(ME)³\cite{17} but restrictions apply to the availability of these data, which were used under license for the current study, and so are not publicly available.Data are however available from the authors upon reasonable request and with permission of CASME II\cite{37}, SAMM\cite{38}, SMIC\cite{39} and CAS(ME)³\cite{17}.\\
\textbf{Materials availability} Not applicable.\\
\textbf{Code availability} If code is needed, please contact the author via email at liruolin918@gmail.com.\\
\textbf{Author Contribution}
Ruolin Li: Data curation, Investigation, Methodology, Software, Visualization, Writing – original draft.
Lu Wang:Conceptualization, Methodology, Supervision, Writing – review \& editing.
Tingting Yang: Validation, Visualization, Writing – original draft, Writing – review \& editing.
Lisheng Xu: Project administration, Supervision, Writing – review \& editing.
Bingyang Ma: Validation, Writing – review \& editing.
Yongchun Li: Funding acquisition, Supervision.
Hongchao Wei: Funding acquisition, Supervision.

\bibliography{MoExt}

{\includegraphics[width=1in,height=1.25in,clip,keepaspectratio]{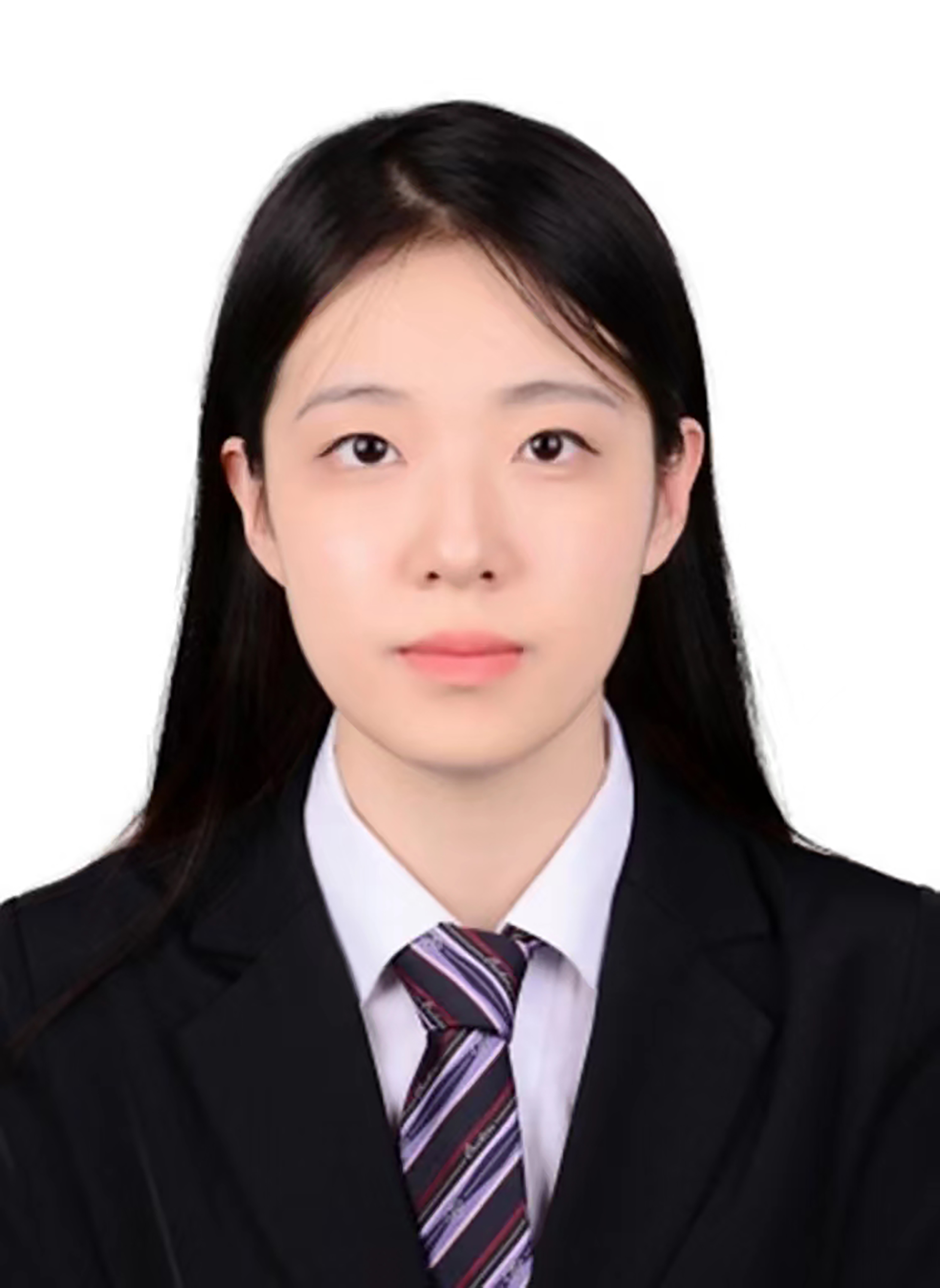}}
\textbf{Ruolin Li} received the B.S. degree from Northeast Forestry University, Harbin, China, in 2021, and she is currently pursuing a Master's degree in Computer Science and Technology at Northeastern University, Shenyang, China. Her research focuses on computer vision-based micro-expression recognition.\par
{\includegraphics[width=1in,height=1.25in,clip,keepaspectratio]{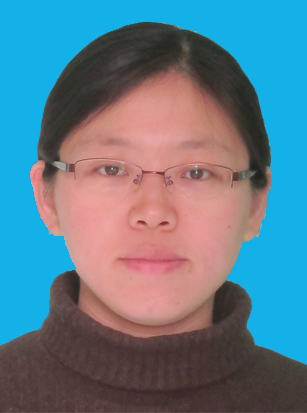}}
\textbf{Lu Wang} received the B.Eng. and M.Eng. degrees in computer science from Harbin Institute of Technology, Harbin, China, in 2003 and 2005, and the Ph.D. degree in electronic engineering from The University of Hong Kong, Hong Kong, China, in 2011. She is currently an Associate Professor with the School of Computer Science and Engineering, Northeastern University, Shenyang, China. Her research interests include computer vision, pattern recognition and biomedical signal and image analysis.\par
{\includegraphics[width=1in,height=1.25in,clip,keepaspectratio]{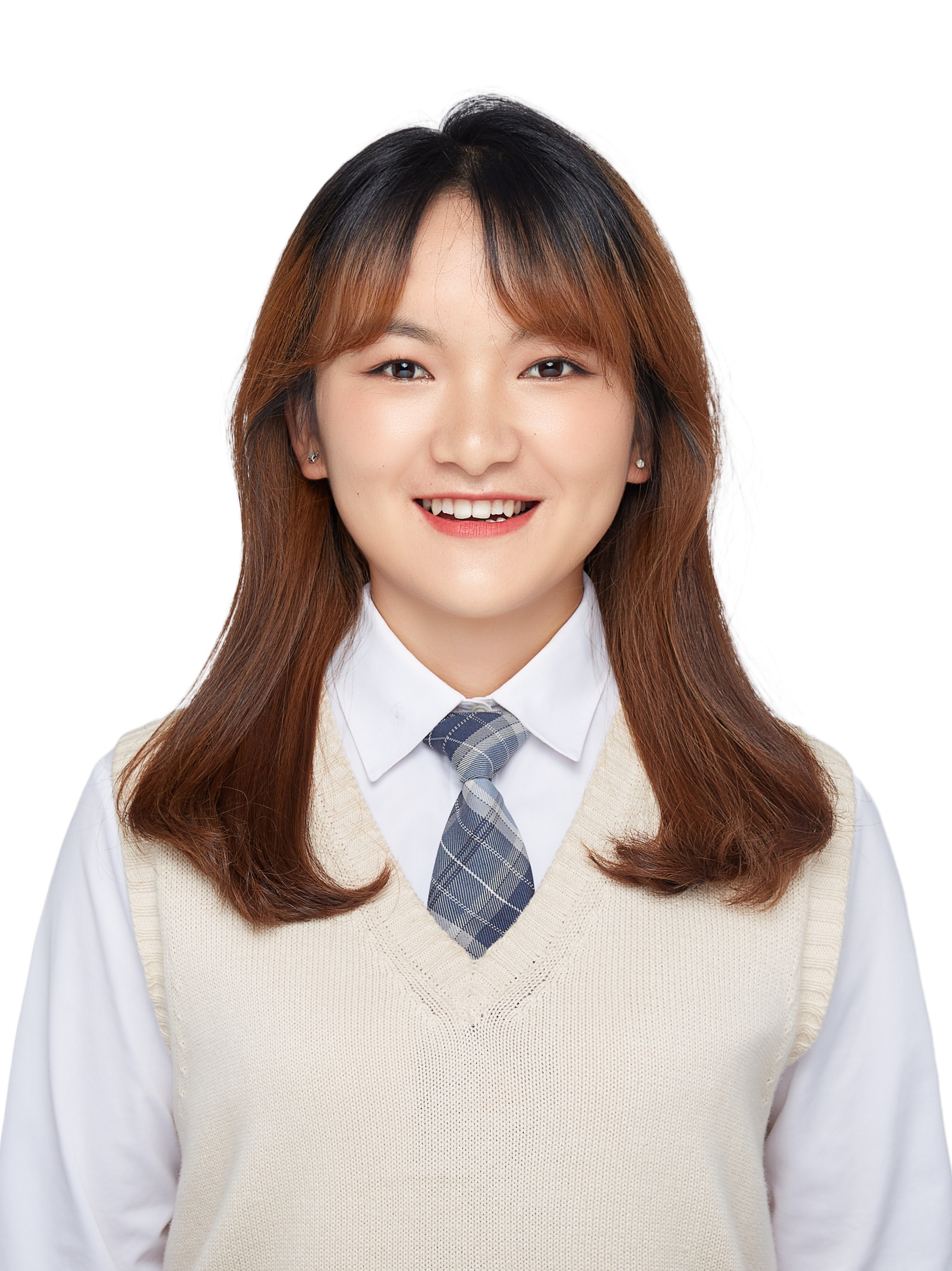}}
\textbf{Tingting Yang} received the B.S. degree from the School of Computer Science and Engineering, Northeastern University, China, in 2022. She is currently working toward the M.S. degree at Northeastern University, Shenyang, China. Her current research interests include computer vision and micro-expression.\par
{\includegraphics[width=1in,height=1.25in,clip,keepaspectratio]{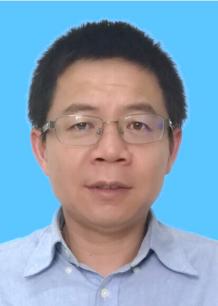}}
\textbf{Lisheng Xu} (SM’15) received the B.S. degree in electrical power system automation, the M.S. degree in mechanical electronics, and the Ph.D. degree in computer science and technology from Harbin Institute of Technology, Harbin, China, in 1998, 2000, and 2006, respectively. He is currently a full professor in the College of Medicine and Biological Information Engineering, Northeastern University, China. He has authored or coauthored more than 200 international research papers, and holds 21 patents and 3 pending patents. His current research interests include nonlinear medical signal processing, computational electromagnetic simulation, medical imaging, and pattern recognition. Prof. Xu is the director of theory and education professional committee of China Medical Informatics Association. He is senior member of IEEE and Chinese Society of Biomedical Engineering. He is the member of the editor board for many international journals such as Physiological Measurement, Biomedical Engineering Online, Computers in Biology and Medicine and so on.\par
{\includegraphics[width=1in,height=1.25in,clip,keepaspectratio]{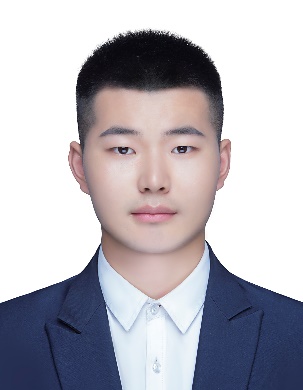}}
\textbf{Bingyang Ma} received the B.S. degree in Software Engineering from Lanzhou University of Technology, Lanzhou, China, in 2021. He is currently working toward the M.S. degree in Computer Science and Technology with the School of Computer Science and Engineering, Northeastern University, Shenyang, China. His research interests include Artificial Intelligence and Micro-expression Recognition.\par
{\includegraphics[width=1in,height=1.25in,clip,keepaspectratio]{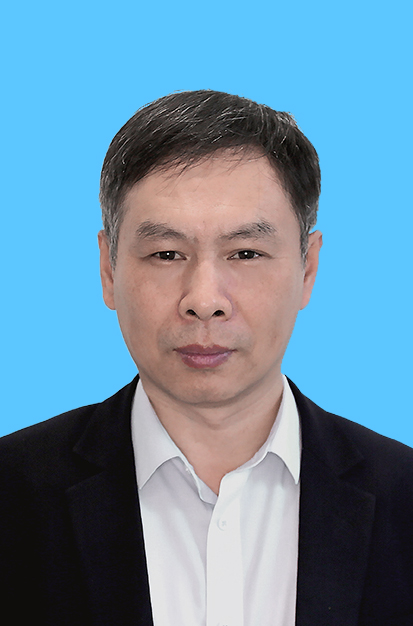}}
\textbf{Yongchun Li} (SM’15) received the B.S. degree in computer and application, and also has obtained Liaoning Province second prize of scientific and technological progress, Shenyang high-level leading talent and other honors. Now is the product director of Shenyang Contain Electronic Technology Co., Ltd. and holds 10 patents and 5 pending patents. Current research interests include engineering management of computer software, physiological signal processing technology for health, machine perception of emotion, emotion expression, utilization technology of emotion, etc.\par
{\includegraphics[width=1in,height=1.25in,clip,keepaspectratio]{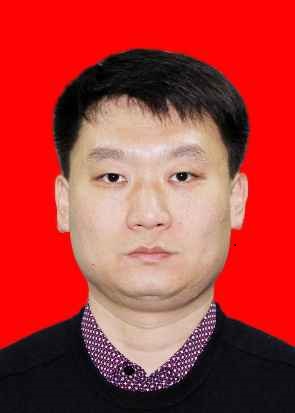}}
\textbf{Hongchao Wei} received the M.S. degree in Economics and National Economy from the Belarusian State Agricultural and Technical University. Now is an adjunct professor at Liaoning University of Science and Technology and Shenyang City University. He has presided over 17 provincial and municipal projects and published 23 academic papers (including 6 Russian papers and 4 English papers). He is a member of the National Democratic Construction Committee of China. He is Liaoning Hundred million Talents Project - thousand level talents, Shenyang top-notch talents. In 2021, He won one third prize of Science and Technology Progress of Liaoning Province.

\end{document}